\documentclass{article} 

\usepackage[pdftex]{graphicx} 

\usepackage{hyperref}
\usepackage{url}

\title{Geometry of Deep Convolutional Networks}

\author{Stefan Carlsson \\ 
School of EECS\\
KTH\\
Stockholm, Sweden \\
} 

\begin{document}

\maketitle

\begin{abstract}
  We give a formal procedure for computing preimages of convolutional
  network outputs using the dual basis defined from the set of
  hyperplanes associated with the layers of the network. We point out
  the special symmetry associated with arrangements of hyperplanes of
  convolutional networks that take the form of regular
  multidimensional polyhedral cones. We discuss  the efficiency of 
  large number of layers of nested cones that result from incremental
  small size convolutions in order to give a good compromise between
  efficient contraction of data to low dimensions and shaping of
  preimage manifolds. We demonstrate how a specific network flattens a
  non linear input manifold to an affine output manifold and discuss
  its relevance to understanding classification properties of deep
  networks.
\end{abstract}

\section{Introduction}

Deep convolutional networks for classification map input data domains
to output domains that ideally correspond to various classes. The
ability of deep networks to construct various mappings has been the
subject of several studies over the years
\cite{arora2016understanding, bengio2011expressive,
  montufar2014number} and in general resulted in various estimates of
capacity given a network structure.  The actual mappings that are
learnt by training a specific network however, often raise a set of
questions such as why are increasingly deeper networks advantageous
\cite{simonyan2014very, szegedy2015going} ?  What are the mechanisms
responsible for the successful generalisation properties of deep
networks ? Also the basic question why deep learning over large
datasets is so much more effective than earlier machine learning
approaches is still essentially open, \cite{krizhevsky2012imagenet}.
These questions are not in general answered by studies of capacity. A
more direct approach based on actual trained networks and the mappings
they are efficiently able to produce seems needed in order to answer
these questions.  It seems ever more likely e.g that the ability of
deep networks to generalize is connected with some sort of restriction
of mappings that they theoretically can produce and that these
mappings are ideally adapted to the problem for which deep learning
has proven successful, Due to the complexity of deep networks the
actual computation of how input domains  are mapped to output
classifiers has been considered prohibitively difficult. From general
considerations of networks with rectifier (ReLU) non linearities we
know that these functions must be piecewise linear
\cite{montufar2014number} but the relation between network parameters
such has convolutional filter weights and fully connected layer
parameters and the  actual functions remains largely obscure. In
general, work has therefore been concentrated on empirical studies of
actual trained networks \cite{dosovitskiy2016inverting,
  mahendran15understanding, mahendran16visualizing}


Recently however there have been attempts to understand the relation
between networks and their mapping properties from a more general and
theoretical point of view. This has included specific procedures for
generating preimages of network outputs \cite{carlsson2017preimage}
and more systematic studies of the nature of piecewise linear
functions and mappings involved in deep networks,
\cite{BasriJ16,Balestrierospline, zhangtropical}.

In this work we will make the assertion that understanding the
geometry of deep networks and the manifolds of data they process is 
an effective way to understand the comparative success of deep
networks. We will consider convolutional networks with ReLU non
linearities. These can be completely characterised by the
corresponding hyperplanes associated with individual convolutional
kernels .  We will demonstrate that the individual arrangement of
hyperplanes inside a layer and the relative arrangement between layers
is crucial to the understanding the success of various deep network
structures and how they map data from input domains to output classifiers.

We will consider only the convolutional part of a deep network
with a single channel. We will assume no subsampling or max pooling. This will
allow us to get a clear understanding of the role of the convolutional part.
A more complete analysis involving multiple channels and fully connected 
layers is possible but more complex and will be left to future work. 

The focus of our study is to analyse how domains of input data are
mapped through a deep network. A complete understanding of this
mapping and its inverse or preimage will give a detailed description
of the workings of the network. Since we are not considering the final
fully connected layers we will demonstrate how to compute in detail
the structure of input data manifold that can be mapped to a specified
reduced dimensionality affine manifold in the activity space of the
final convolutional output layer. This flattening of input data is
often considered as a necessary preprocessing step for efficient
classification. 

The understanding of mappings between layers will be based on the
specific understanding of how to compute preimages for networks
activities. We will recapitulate and extend the work in
\cite{carlsson2017preimage} based on the construction of a dual basis from an
arrangement of hyperplanes. By specialising to convolutional networks
we will demonstrate that the arrangement of hyperplanes associated
with a specific layer can be effectively described by a
regular multidimensional polyhedral cone oriented in the identity direction in
the input space of the layer. Cones associated with successive layers
are then in general partly nested inside their predecessor. This leads
to efficient contraction and shaping of the input domain data
manifold.  In general however contraction and shaping are in conflict
in the sense that efficient contraction implies less efficient
shaping. We will argue that this' conflict is resolved by extending
the number of layers of the network with small incremental updates of
filters at each layer.

The main contribution of the paper is the exploitation of the
properties of nested cones in order to explain how non linear
manifolds can be shaped and contracted in order to comply with the
distribution of actual class manifolds and to enable efficient
preprocessing for the final classifier stages of the network. We will
specifically demonstrate the capability of the convolutional part of
the network to flatten non linear input manifolds which has previously
been suggested as an important preprocessing step in object recognition,
\cite{dicarlo2007untangling, rowes2000nonlinear}

\section{Layer mappings in ReLU networks and their preimage}

Transformations between layers in a network  with ReLU as nonlinear elements
can be written as
\begin{equation}
y \ = \ [{\bf W} x + b\  ]_+ 
\end{equation}
Where $ [\ \ ]_+ $ denotes the ReLU function $max(0,x_i)$.  applied
component wise to the elements of the input vector $x$ which will be
confined to the positive orthant of the d-dimensional Euclidean input
space. It divides the components of the output vector $y$
into two classes depending on the location of the input $x$:
\begin{eqnarray}
\{j: &w_j^{T} x + b_j    > 0\} &\rightarrow \ \ \ \ y_j = w_j^{T} x + b_j  \cr\cr
\{i: &w_i^{T} x + b_i \leq 0 \}&\rightarrow \ \ \ \ y_i = 0 
\end{eqnarray}
In order to analyse the way domains are mapped through the network we will
be interested in the set of inputs $x$ that can generate a specific output $y$.
\begin{eqnarray}
P(y) = \{ x: y  = [{\bf W} x + b\  ]_+ \} 
\end{eqnarray}
This set, known as the preimage of $y$ can be empty, contain a unique
element $x$ or consist of a whole domain of the input space. This last
case is quite obvious by considering the ReLU nonlinearity that maps
whole half spaces of the input domain to $0$ components of the output $y$. 

The preimage will depend on the location of the input relative to the
arrangement of the hyperplanes defined by the affine part of the
mapping:
\begin{eqnarray}
  \Pi_i = &\{x: w_i^{T} x + b_i = 0 \}  \ \ \ \ \ \   i = 1, 2, \ldots d
\end{eqnarray}
These hyperplanes divides the input space into a maximum of $2^d$
number of different cells with the maximum attained if all hyperplanes
cut through the input space which we take as the non negative orthant of
the d-dimensional Euclidean space $R_{+}^d$. Understanding the
arrangement of these hyperplanes in general and especially in the case
of convolutional mappings will be central to our understanding of how
input domains are contracted and collapsed through the network.

The preimage problem can be treated geometrically using these
hyperplanes as well as the constraint input domains defined by
these. For a given output $y$  we can denote the components where
$y_j > 0$ as $y_{j_1}, y_{j_1} \ldots y_{j_q}$ and the complementary
index set where $y_i = 0$ as ${i_1}, {i_1} \ldots {i_p}$ With each positive
component of $y$ we can associate a hyperplane:
\begin{eqnarray}
\Pi_j^* = \{x:  y_j = w_j^{T} x + b_j \} \ \ \ \ \ \ j = j_1, j_2,  \ldots j_q   
\end{eqnarray}
which is just the hyperplane $\Pi_j$ translated with the output $y_j$
For the $0$-components of $y$ we can define the half spaces
\begin{eqnarray}
X_i^- \ = \{x: w_i^{T} x + b_i \leq 0 \} \ \ \ \ \ \ i = i_1, i_2,  \ldots i_p   
\end{eqnarray}
I.e the half space cut out by the negative side of the plane $\Pi_i$.
These planes and half spaces together with the general input domain constraint of 
being inside $R_d^+$ define the preimage constraints given the output $y$. \newline
If we define the affine intersection subspace:
\begin{eqnarray}
 \Pi^*\ =  \Pi^*_{j_1}   \cap \Pi^*_{j_2}  \cap \ldots \cap \Pi^*_{j_q}  
\end{eqnarray}
and the intersection of half spaces:
\begin{eqnarray}
X^-  \ =   X^-_{i_1} \cap X^-_{i_2} \cap  \ldots \cap X^-_{i_p}    
\end{eqnarray}
the preimage of $y$ can be defined as:
\begin{eqnarray}
P(y) \ =   \Pi^* \cap X^- \cap R^d_+   
\end{eqnarray}

The constraint sets and the preimage set is illustrated in figure
\ref{fig1} for the case of $d=3$ and various outputs $y$ with
different number of $0$-components. 

For fully connected networks, computing the preimage set amounts to
finding the intersection of an affine subspace with a polytope in $d-$
dimensional space. This problem is known to be exponential in $d$ and
therefore intractable. However, we will see that this situation is
changed substantially when we consider convolutional instead of fully
connected networks.

\section{The dual basis for expressing the preimage}

In order to get more insight into the nature of preimages we will
devise a general method of computing that highlights the nature of the
arrangement of hyperplanes.
The set of hyperplanes $\Pi_i,\  i = 1 \ \ \ \ldots d$ will be assumed to
be in general position, i.e. no two planes are parallel. The
intersection of all hyperplanes excluding plane $i$:
\begin{eqnarray}
S_i \ = \ \Pi_1   \cap \Pi_2  \cap \ldots \cap \Pi_{i-1}   \cap \Pi_{i+1}  \ldots \cap \Pi_d    
\end{eqnarray}
is then a one-dimensional affine subspace $S_i$ that is contained in
all hyperplanes $\Pi_j$ excluding j = i. For all $i$ we can define
vectors $ e_i$ in $ R^d$ parallel to $ S_i$. The general position of
the hyperplanes then guarantees that the set $ e_i$ is complete in
$ R^d$. By translating all vectors $ e_i$ to the point in $ R^d$ which is the
mutual intersection of all planes $ \Pi_i $
\begin{eqnarray}
\Pi_1   \cap \Pi_2  \cap \ldots  \cap \Pi_d    
\end{eqnarray}
they can therefore be used as  a basis that spans $ R^d$.
This construction also  has the property that  the intersection of the subset of hyperplanes:
\begin{eqnarray}
\Pi_{j_1}   \cap \Pi_{j_2}  \cap   \ldots \cap \Pi{j_q}
\end{eqnarray}
is spanned by the complementary dual basis set 
\begin{eqnarray}
e_{i_1} , e_{i_2}  \ldots  e_{i_p}    
\end{eqnarray}
The dual basis can now be used to express the solution to the preimage
problem.  The affine intersection subspace $ P^*$ associated with the
positive components $ j_1 j_2 \ldots j_q$ of the output $y$ is spanned
by the complementary vectors associated with the negative components
$ i_1 i_2 \ldots i_p $ . These indices also define the hyperplanes
$ \Pi_1, \Pi_2 \ldots \Pi_p $ that constrain the preimage to lie
in the intersections of half spaces associated with the negative sides.

\begin{figure}[h]
\begin{center}
\includegraphics[width=0.35\linewidth]{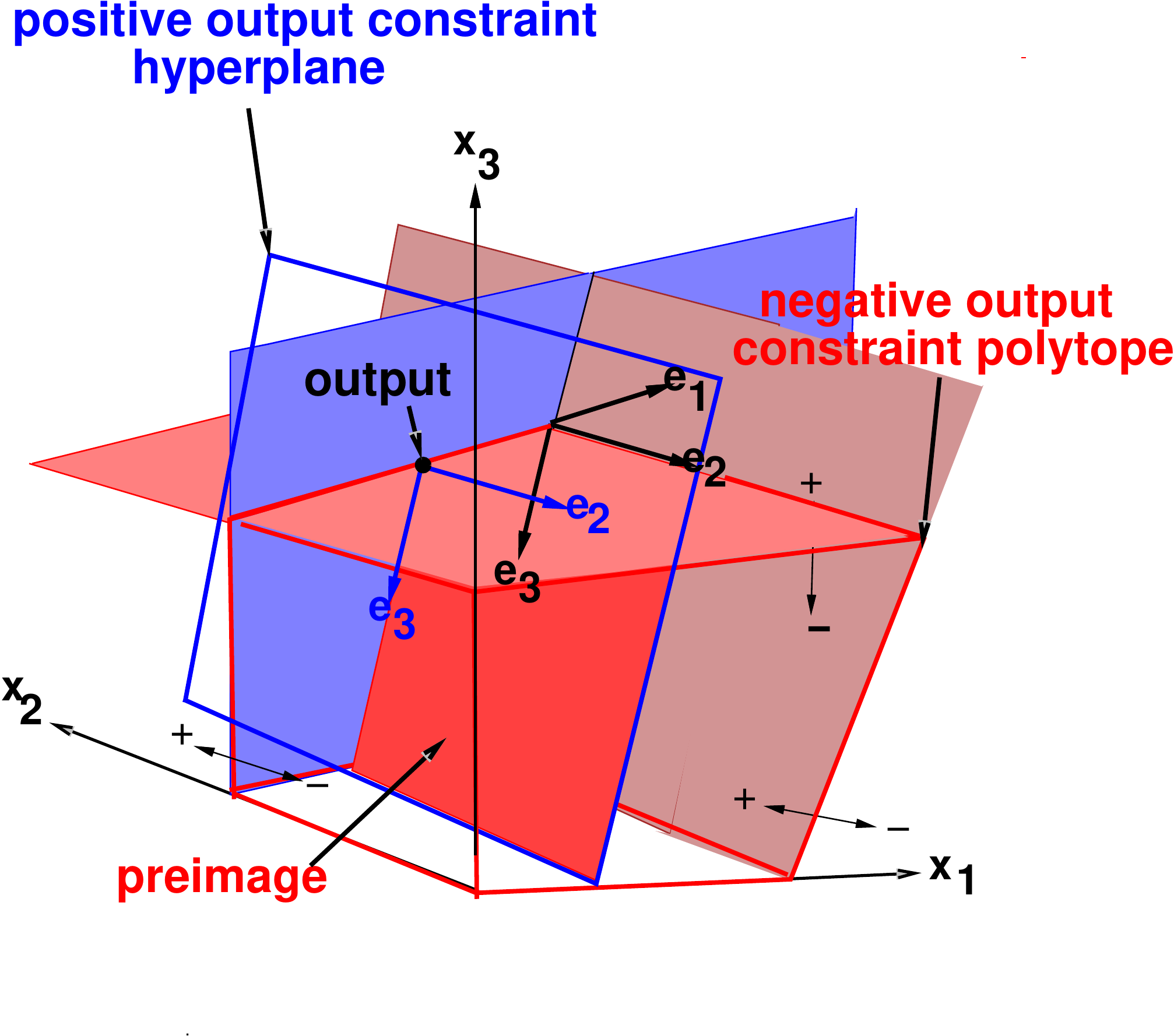}
\includegraphics[width=0.60\linewidth]{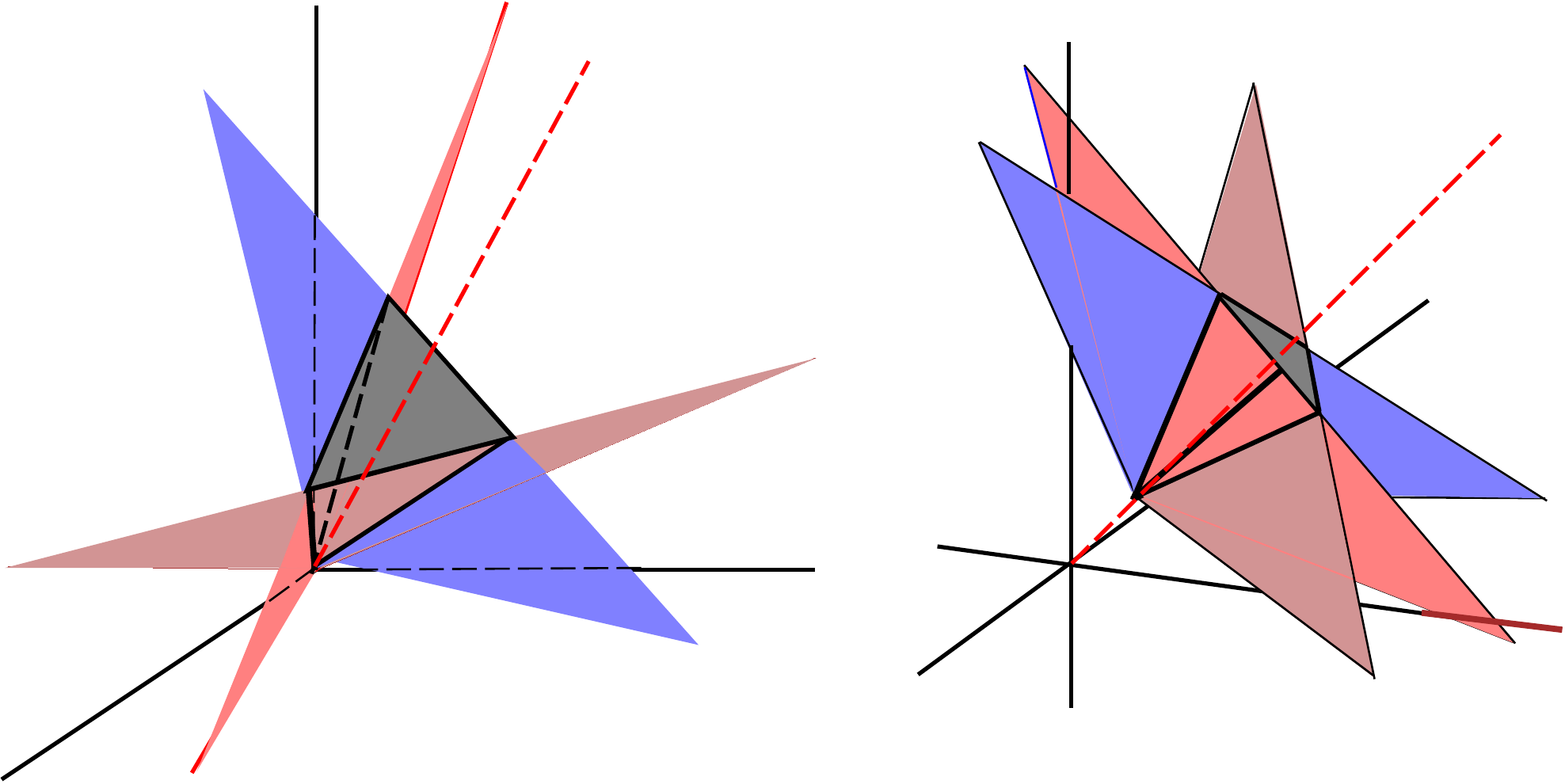}
\label{fig1}
\end{center}
\caption{\it  {\bf Left}: 3 planes in general position and the preimage (red) of the output 
(black dot)  $e_2$ and $e_3$ components of dual basis used to generate preimage. {\bf Right}:
Polyhedral cone of 3 planes from a circulant layer transformation matrix. (two different views)
The nesting  property of the cone will refer to it's ability to ``grip'' the coordinate axis
of the input space.}
\end{figure}

We now define the {\it positive} direction of the vector $ e_i$ as
that associated with the {\it negative} side of the plane $ \Pi_i$.
If we consider the intersection of the subspace $ P^*$ and the
subspace generated by the intersections of the hyperplanes $ \Pi_i$
associated with the negative components of $y$ we get:
\begin{eqnarray}
 \Pi^*_{j_1}   \cap \Pi^*_{j_2}  \cap \ldots \cap \Pi^*_{j_q}  \cap \Pi_{i_1}  \cap \ldots \cap \Pi_{i_p}  
\end{eqnarray}
Due to complementarity of the positive and negative indices, this is a
unique element $x^* \in  R^d$ (marked ``output'' in figure \ref{fig1} which
lies in the affine subspace of the positive output components $ P^*$
as well as on the intersection of the boundary hyperplanes $ \Pi_i$
that make up the half space intersection constraint $ X^-$ for the
preimage. if we take the subset of the dual basis vectors with
$e_{i_1} , e_{i_2} \ldots e_{i_p}$ and move them to this intersection
element, they will span the part of the negative constraint region
$ X^-$ associated with the preimage. I.e. the preimage of the output
$ y$ is given by:
\begin{eqnarray}
P(y) = \{ x \in R^d_+ : \  \  \ x =  x^* + \sum_{i=1}^{i=i_p} \alpha_i e_i \ \ \ \  \alpha_i \geq 0 \}
\end{eqnarray}

\section{Arrangements of hyperplanes for convolutional layers form regular polyhedral cones}


We will now specialise to the standard case of convolutional networks.
In order to emphasize the basic role of geometric properties we will
consider only a single channel with no subsampling.  Most of what we
state will generalize to the more general case of multiple channels
with different convolutional kernels but needs a more careful analysis
we will exploit the fact that convolutional matrices are in most
respects asymptotically equivalent to those of circulant matrices
where each new row is a one element cyclic shift of the previous.  For
any convolution matrix we will consider the corresponding circulant
matrix that appends rows at the end to make it square and circulant.
Especially when the support of the convolution is small relative to
the dimension d, typically the order of 10 in relation to 1000, this
approximation will be negligible.  Except for special cases the
corresponding circulant will be full rank, which means that properties
about dual basis etc. derived previously will apply also here. As is
standard we will assume that the bias b is the same for all
applications of the convolution kernels.

The first thing to note about hyperplanes associated with circulant
matrices is that they all intersect on the identity line going through
the origin and the point $ (1,1, \ldots 1)$ . Denote the circulant
matrix as $C$ with elements $c_{i,j}$. The circulant property implies
$c_{i+1,j} = c_{i,j-1}, \ \ i = 1 \ldots d-1, \ \ j = 2 \ldots d$ and
$c_{i+1,1} = c_{i,d}$.  Each row is shifted one step cyclically relative
to the previous. For the hyperplane corresponding to row i we have:
\begin{eqnarray}
\sum_{j=1}^{j=d} c_{i,j} x_j \ + b = 0 
\end{eqnarray}
It is easy to see that the circulant property implies that the sum of all
elements along a row is the same for all rows.  Let the sum of the row be
$a$. We then get: $x_j = -b/a$ for $j = 1 \ldots d$ as a solution for
this system of equations which is a point on the identity line in $R^d$.

The arrangement of the set of hyperplanes:
\begin{eqnarray}
w_i^{T} x + b = 0  \ \ \ \ \ \ i = 1 \ldots d  
\end{eqnarray}
with $w_i^{T}$ the $i$:th row of the circulant augmented convolutional
matrix $W$, will be highly regular. Consider a cyclic permutation $Px$ of
the components of the input $x$ described by the single shift matrix $P$
i.e $x_i$ is mapped to $x_{i+1}$ for $i = 1 \ldots d-1$ and $x_d$ is mapped 
to $x_1$. We then get:
\begin{eqnarray}
w_i^{T} Px + b&= \ w_{i+1}^{T} x + b&= \ 0  \ \ \ \ \ \ i = 1 \ldots d-1  \cr \cr
w_d^{T} Px + b&= \ w_1^{T} x + b&= \ 0  
\end{eqnarray}
which states that points on the hyperplane associated with weights
$w_i$ are mapped to hyperplane associated with weights $w_{i+1}$. The
hyperplanes associated with the weights $w_i$ $i = 1 \ldots d$
therefore form a regular multidimensional polyhedral cone in $R^d$ around the identity
line, with the apex located at $x^T = (-b/a, -b/a \ldots-b/a)$
controlled by the bias $b$ and the sum of filter weights
$a$. Geometrically, the cone is determined by the apex location, the
angle of the planes to the central identity line and its rotation in
$d$-dimensional space. Apex location and angle are two parameters
which leaves $d-2$ parameters for the multidimensional rotation
in $R^d$. This maximum degree of freedom is however attained only for
unrestricted circulant transformations. The finite support of the
convolution weights in CNN:s will heavily restrict rotations of the
cone. The implications of this will be discussed later.

\section{Nested cones efficiently contract input data}

Any transformation between two layers in a convolutional network can
now be considered as a mapping between two regular multidimensional
polyhedral cones that are symmetric around the identity line in
$R^d$. The coordinate planes of the input space $R^d_+$ can be modelled as
such a cone as well as the output space given by the convolution. The
strong regularity of these cones will of course impose strong
regularities on the geometric description of the mapping between
layers. Just as in the general case, this transformation will be
broken down to transformations between intersection subspaces of the
two cones.

\begin{figure}[h]
\begin{center}
\includegraphics[width=0.75\linewidth]{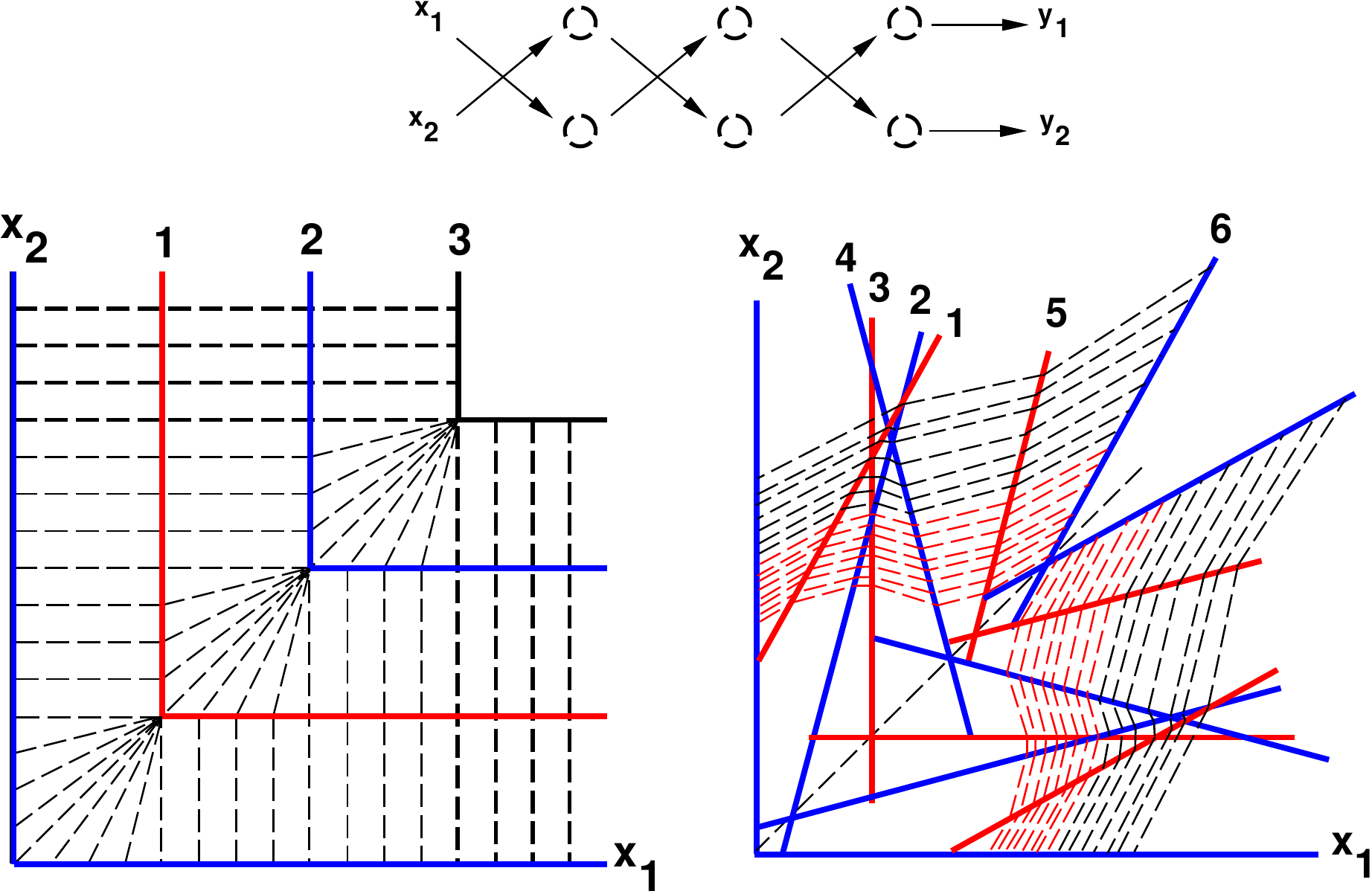}
\label{fig2}
\end{center}
\caption{\it Illustration of how data in 2d input space $(x_1, x_2)$ contracts
by successive layers in multi layer 2 node networks with circulant transformations. 
{\bf Left}: Transformation $(a,b) = (1,0)$  . Only the bias differs between layers. Alternating
red and blue frames show successive layers remapped to the input space. {\bf Right}: Arbitrary
circulant transformations. Note how the nesting property insures  a good variation in the
generated manifolds. When nesting becomes less pronounced for higher values of input
the variation of the manifolds diminishes.} 
\end{figure}

In order to get an idea of this we will start with a simple multi
layer network with two dimensional input and output and a circulant
transformation:
\begin{eqnarray}
  x^{(l+1)}_1 &= [a^{(l)} x^{(l)}_1 + b^{(l)} x^{(l)}_2 + c^{(l)}]_+\cr \cr
  x^{(l+1)}_2 &= [b^{(l)} x^{(l)}_1 + a^{(l)} x^{(l)}_2 + c^{(l)}]_+ 
\end{eqnarray}
Figure \ref{fig2}  illustrates the mapping of data from the input space
$(x_1, x_2)$ to the output space $(y_1, y_2)$ for two networks with 3 and
6 layers respectively.
The dashed lines represent successive preimages of data that maps to a
specific location at a layer. By connecting them we get domains of
input data mapped to the same output at the final layer, i.e they are
contraction flows depicting how data is moved through the network. Note
that in both layers the major part of the input domain is mapped to
output $(0, 0)$. This is illustrated for the first trivial bias only
layer with $a=1, b=0$. The domain of the input that is mapped to output
domain is just quite trivial planar manifolds.

The second network with more varied weights illustrates how input
domain manifolds with more structure can be created. It also
demonstrates the importance of the concept of ``nested cones'' and how
this affects the input data manifolds. The red lines represent data
that is associated with layer cones that are completely nested inside
its predecessors, while the black lines represent data where the
succeeding cone has a wider angle than its predecessor. When this
happens, the hyperplanes associated with the output cone will
intersect the hyperplanes of the input cone and input data beyond this
intersection is just transformed linearly. Since all data in figure
\ref{fig2} is remapped to the input space this has the effect that data is
not transformed at all. This has no effect at all at the shaping of the
input manifold. One could say that these layers are ``wasted'' beyond
the location of the intersection as far as network properties are
concerned since they neither contribute to the shaping or the
contraction of input data manifolds. The effect of this on the input
manifold can be seen as a less diverse variation of its shape (black)
compared to the previous part associated with the completely nested
part of the layer cones.

In higher dimensions the effects of nested vs. partially nested cones
appear in the same way but more elaborate. In addition to the 2d case
we also have to consider rotations of the cone, which as was pointed
out earlier, has $d-2$ degrees of freedom for cones in $d$ dimensional
space. The effects of contraction of data from higher to lower
dimensions also become more intricate as the number of different
subspace dimensionalities increases. Most of these effects can be
illustrated with 3 dimensional input and output spaces. For $d=3$ the
generic circulant matrix can be used to define a layer transformation:
\begin{eqnarray}
        y_1 &= [a x_1 + b x_2 + c x_3 + d]_+ \cr 
        y_2 &= [c x_1 + a x_2 + b x_3 + d]_+ \cr 
        y_3 &= [b x_1 + c x_2 + a x_3 + d]_+ 
\end{eqnarray}
The transformation properties of this network are most easily
illustrated if we start with the pure bias case with transformation
$W=I$, i.e $a=1, b=0,  c=0$. A specific element in input space is mapped
according to its position relative to the hyperplanes. If we use the
dual basis to define the coordinates of the output data, the mapping
for the input element will be the same in input cells with the same
relation to all hyperplanes.  In $d$ dimensions, the hyperplanes
divide the input space into $2^d$ cells where elements are mapped to a
specific associated intersection subspace in the output domain. 

\begin{figure}[h]
\begin{center}
\includegraphics[width=1.0\linewidth]{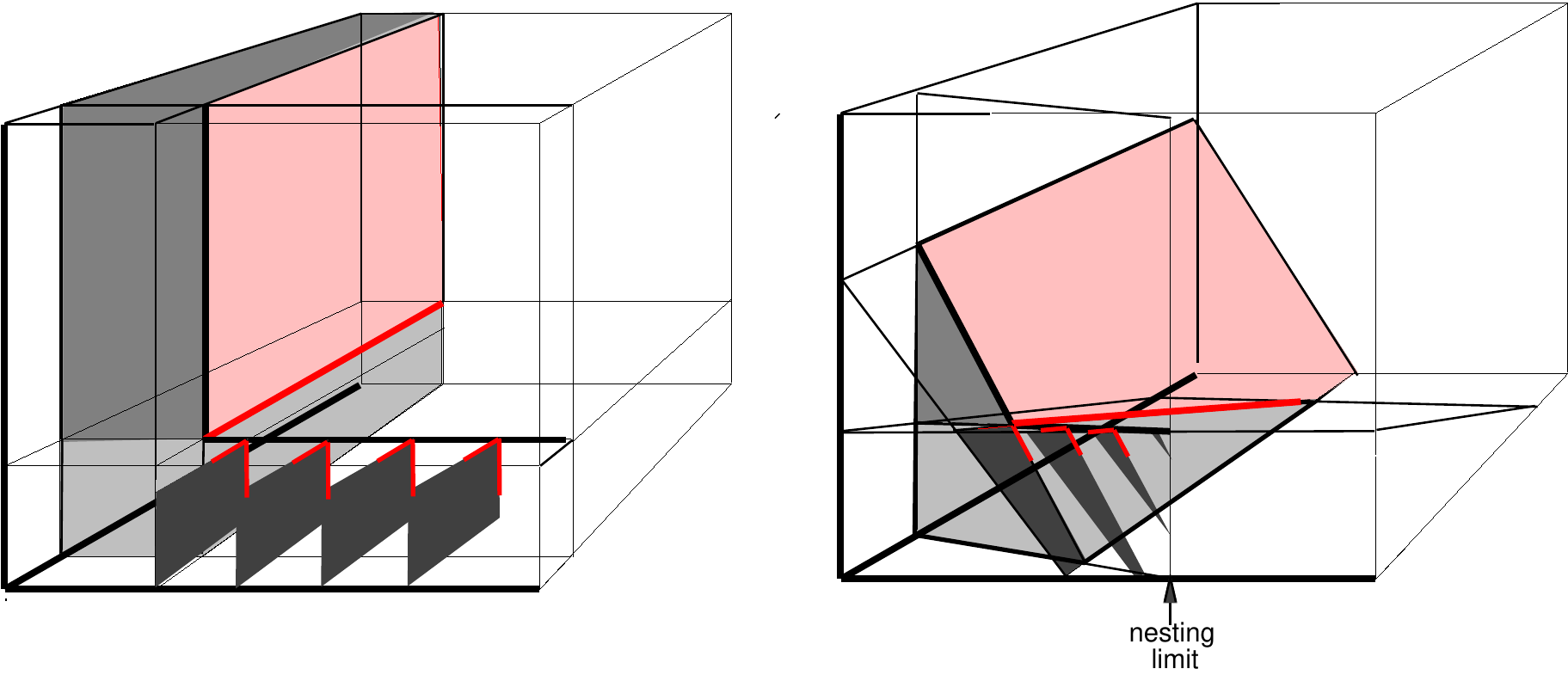}
\label{fig3}
\end{center}
\caption{\it Illustration of how data maps between  intersection subspaces in two successive
layers of 3 node networks with circulant transformations  {\bf Left}: Identity transformation
and bias only. An element in the input space maps according to it's cell location determined
by the sign relative to the output hyperplanes. The figure illustrates how two different
cells, dark grey and light grey are mapped to red 2d plane and red 1d line in the output
layer. Note that both  3d volume, 2d faces and 1d edges maps to the same output 
intersection subspace. This means that dimension of subspace location for the element
is non increasing which implies that data gets contracted. This is a consequence of the
nesting properties of the output and input polyhedral cones. {\bf Right} The transformation is
now such that the angle of the cone is increased. The planes of the output layer now intersects
 the coordinate axis of the input space. Beyond the intersections the non increasing 
contraction property ceases. The dark gray areas indicate preimages of points located
on the output 1-d coordinate axis. Note how they are decreases in size between the left and
right examples. } 
\end{figure}
The grey shaded boxes indicate two cells with different numbers of
negative constraints 1 and 2 respectively. The content of the upper
one with one negative constraint including all its bounding faces and
their intersections is mapped to a specific $2d$ hyperplane in the
output domain while the content of the lower one with two negative
constraints is mapped to the 1d intersection of two hyperplanes. This
illustrates the most important property of the nesting of the cones
associated with the input and output layer: For a range of
transformations in the vicinity of the identity mapping, the input
space, properly normalised in range, is divided into cells where the
elements of the cells including their bounding faces an their
intersections are mapped to output intersection subspaces with equal
or lower dimension. This means that the content of the cell is
irreversibly contracted to lower dimensions.

Figure \ref{fig3} also contains examples of preimages to individual
elements (dark shaded grey rectangles) and the components of the dual bases
used to span these. Note that these are affected by changing the angle of 
the output cone. It introduces a limit of the nesting beyond which the mapping 
properties of the transformation are changed so that data no longer maps to
a manifold of equal or lower dimensionality. I.e the contraction property
is lost in those regions of the input space where nesting of cones ceases.

We will formally define this important property of nested cones as:
\begin{quote}
  Let $R^d_+$ be the non negative orthant of the Euclidean
  $d$-space. Let $\Pi^0_i$ be the hyperplane defined as $x_i = 0$ for
  $(x_1 \ldots x_d) \in R^d_+$.  Consider a set of corresponding
  hyperplanes $\Pi_1 \ldots \Pi_d$ in $R^d_+$ associated with a
  circulant matrix.  Take a subset $i_1 \ldots i_p$ of these
  hyperplanes and form the intersection subset:
  $ M_{i_1 \ldots i_p} = \Pi_{i_1} \cap \ldots \cap\Pi_{i_p}$ .  If for
  each $x$ in $M_{i_1 \ldots i_p}$ the positive span
  $S_+(e_{j_1} \ldots e_{j_q})$ of the associated dual basis contains
  an element in the corresponding intersection subset
  $\Pi^0_{i_1} \cap \ldots \cap \Pi^0_{i_p}$ and thereby in each subset
  of planes with these indexes, but no other subset, we say that the
  cone formed by the hyperplanes $\Pi$ is completely nested in the cone
  formed by the hyperplanes $\Pi^0$.
\end{quote}

We see that this definition implies that the cone formed by planes
$\Pi$ is completely contained in that formed by the planes $\Pi^0$ but
also that its relative rotation is restricted. We will have reason to
relax the condition of inclusion of all elements $i_1 \ldots i_p$ in
the intersection subset and talk about cones with restricted
nesting. Complete nesting implies contraction of data from one layer
to the next which can be seen from the fact that all elements of the
complete intersection subset
$\Pi^0_{i_1} \cap \ldots \cap \Pi^0_{i_p}$ and thereby in each subset
are mapped to the intersection subset
$ M_{i_1 \ldots i_p} = \Pi_{i_1} \cap \ldots \cap\Pi_{i_p}$ with same
dimensionality. In addition elements from intersection subsets formed
by subsets of indexes $i_1 \ldots i_p$ will also be mapped to this
same intersection subset. The subsets associated with these indexes
are however of higher dimension.  Consequently, mapping of data
between layers will be from intersection subsets to intersection
subsets with equal or lower dimension. This is the crucial property
connecting degree of nesting with degree of contraction.

By going further through the network to higher layers this contraction
is iterated and data is increasingly concentrated on intersection
subspaces with lower dimension which is reflected by the increased
sparsity of the nodes of the network. The convolutional part of a deep
network can therefore be seen as a component in a metric learning
system where the purpose of the training will be to create domains in
input space associated with different classes that are mapped to
separate low dimensional outputs as a preprocessing for the final
fully connected layers that will make possible efficient separation of
classes.


There is therefore a conflict between the diversity of input manifolds
that contract to low dimensional outputs and the degree of contraction
that can be generated in a convolutional network. The efficient
resolution of this conflict seems to lie in increasing the number of
layers in the network in order to be able to shape diverse input
manifolds but with small incremental convolution filters that retain
the nesting property of the cone in order to preserve the proper
degree of contraction. Empirically, this is exactly what has been
demonstrated to be the most efficient way to increase performance of
deep networks \cite{simonyan2014very, szegedy2015going}.


\section{Mapping nonlinear input  to affine output manifold }

\begin{figure}[h]
\begin{center}
\includegraphics[width=1.0\linewidth]{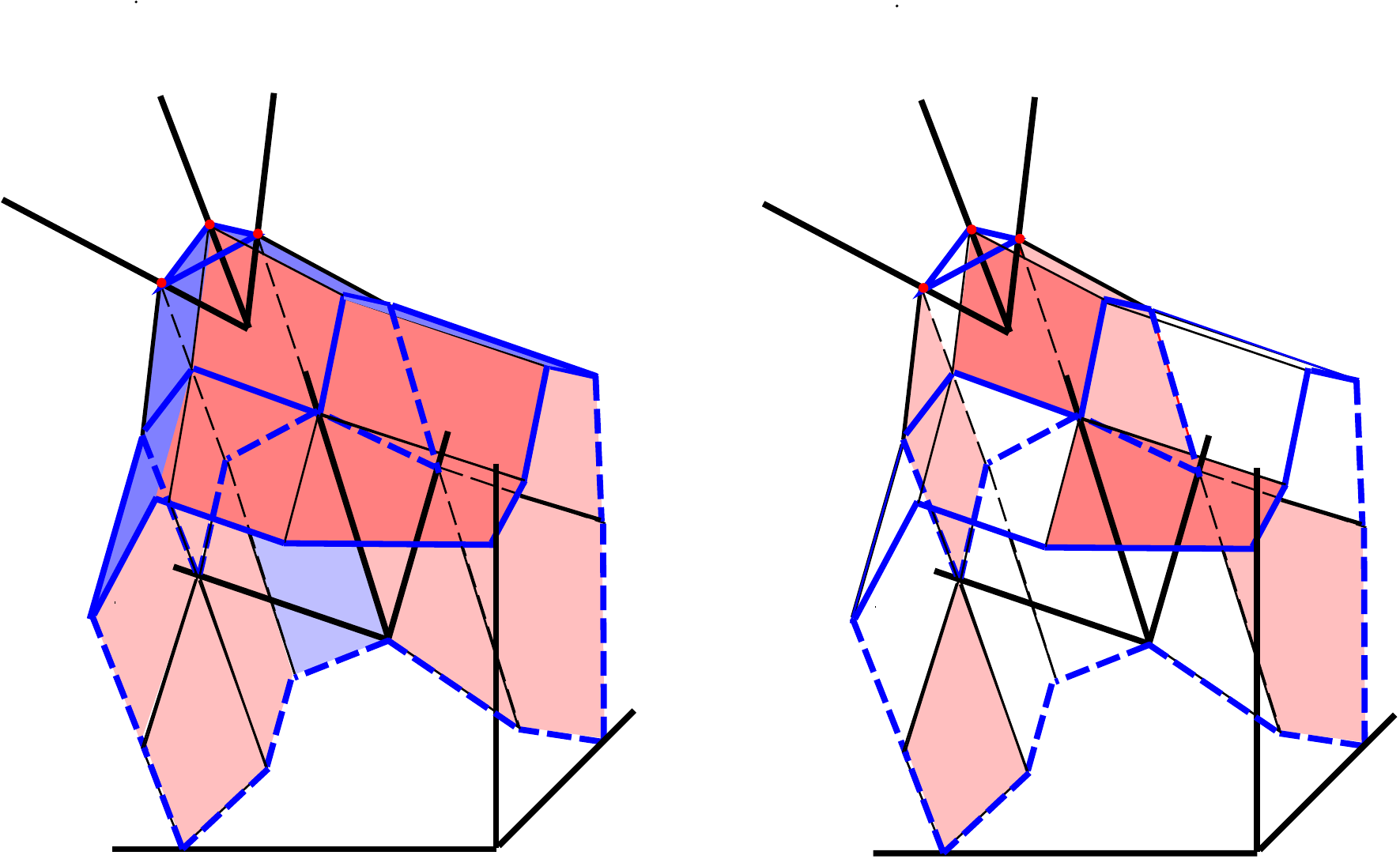}
\label{fig4}
\end{center}
\caption{\it Piecewise planar manifold in 3d input space that maps to
  affine manifold (blue triangle) at the final convolutional layer in
  a 3-node 3-layer network with circulant transformations. All data is
  remapped to the input space. {\bf Left}: Red patches are mapped to 0
  dimensional red points at the three output coordinate axis Blue
  patches are mapped to 1d lines connecting the points. (dark red is
  outside light red is inside of the manifold {\bf Right}: Patches
  that are generated by selective components of the dual basis at each
  layer.  The positive span generated by selective components of the dual basis
  emanating from the red output points on the triangle as well as from
  each intersection with coordinate lines in early layers, intersects 
  with the arrangement of hyperplanes representing the
  preceding layer. The 1-d intersections are then used as seed points for new 
spans that intersect next preceding layer etc. The 2d intersections together 
with selective edges from the spans generate linking patches that ensures the continuity 
of the input manifold. }
as 
\end{figure}

We are now in a position to give a general characterizaton of the
preimage corresponding to a specified output domain at the final
convolutional output layer assuming the property of nested layer
cones. Ideally we would like to include the final fully connected
layers in this analysis but it will require a special study since we
cannot assume the nesting property to be valid for these. In the end
the network should map different classes to linearly separable domains
in order to enable efficient classification.  It is generally
suggested that the preprocessing part of a network corresponds to
flattening nonlinear input manifolds in order to achieve this final
separation at the output. In order to be able to draw as general
conclusions as possible we shall demonstrate the exact structure of a
nonlinear input manifold that maps to a prespecified affine manifold
at the final convolutional layer. We denote this manifold $M$ and the
output at the final convolutional layer by $x^{(l)}$.  The final layer
can be characterised by the set of hyperplanes:
$\Pi^{(l)}_1 \ldots \Pi^{(l)}_d$. Let the zero components of the
output $x^{(l)}$ be $i_1, i_2 \ldots i_q$. It can then be associated
with the intersection of the output manifold and the corresponding
hyperplanes
\begin{eqnarray}
 M \cap \Pi^{(l)}_{i_1}   \cap \Pi^{(l)}_{i_2}  \cap \ldots \cap \Pi^{(l)}_{i_q}  
\end{eqnarray}
The degree of intersection $q$ will depend on the dimensionality of
$M$.  If $M$ is a $d-1$ dimensional hyperplane in general position it
will intersect any combination of hyperplanes at output level
$l$. This is the maximum complexity situation that will generate a
$d-1$ - dimensional input manifold. Reducing dimensionality of $M$
means reducing the possible intersection with combinations of
hyperplanes.  
Note that if $M$
intersects the set the intersection of planes $i_1,
i_2 \ldots
i_p$ it also intersects the intersection of any subset of
these. Intersecting $M$
with each of these subsets will generate pieces of intersections
linked together. These are affine subsets with different
dimensionality and the preimage of each piece will be generated by
complementary dual basis components. This is illustrated by figure
\ref{fig4} for the case of an affine plane in $R^3_+$
intersecting to give a triangular domain.  In this case we have three
points on the coordinate axis, and three lines connecting these.  The
three points will all span 2d planes bases on different pairs of
complementary dual basis components. In addition to these, the points
on the lines of the triangle generated by intersecting $M$
with each of the three individual output planes will generate
$1d$
lines that jointly will span a $2d$
plane.  This plane will connect continuously with the planes spanned
from the points on the axis to yield a piecewise planar input manifold
to the final layer. Continuing through the network, this piecewise
planar manifold will intersect with the planes of layer $l-1$
and the procedure is iterated until we reach the input layer.

This procedure generalises to arbitrary dimensions but the complexity
of course grows with the increasing combinatorics. The basic principle
of layer by layer recursively generating piecewise affine manifolds
still holds.  The complexity lies in the fact that each intersection
of the manifold M with every subset of possible hyperplane
intersections will generate a seeding hyperplane and and each of these
will act as a new manifold $M$ at the next layer. Note however that the
nested cone property substantially reduces complexity compared to the
general case of arbitrary hyperplanes.

%
%
%

%
%

It should be pointed out that these manifold do not necessarily correspond to
actual class manifolds since we are not considering the complete network with
fully connected layers. They can however be considered as more elaborate and 
specific building blocks in order to construct the actual class manifolds of 
a trained network.

\section{Summary and conclusions}

We have defined a formal procedure for computing preimages of deep
linear transformation networks with ReLU non linearities using the
dual basis extracted from the set of hyperplanes representing the
transformation. Specialising to convolutional networks we demonstrate
that the complexity and the symmetry of the arrangement of
corresponding hyperplanes is substantially reduced and we show that
these arrangements can be modelled closely with multidimensisional
regular polyhedral cones around the identity line in input space.  We
point out the crucial property of nested cones which guarantees efficient
contraction of data to lower dimensions and argue that this
property could be relevant in the design of real networks. By
increasing the number of layers to shape input manifolds in the form
of preimages we can retain the nested cone property that most
efficiently exploits network data in order to construct input manifolds
that comply with manifolds corresponding to real classes and would
explain the success of ever deeper networks for deep learning. The
retaining of the nested cone property can be expressed as a limitation
of the degrees of freedom of multidimensional rotation of the
cones. Since convolutional networks essentially always have limited
spatial support convolutions, this is to a high degree built in to
existing systems. The desire  to retain the property of nesting could however
act as an extra constraint to further reduce the complexity of the convolutions.
This of course means that the degrees of freedom are reduced for a network which
could act as a regularization constraint and potentially explain the puzzling
efficiency of generalisation of deep networks in spite of a high number of 
parameters.

We demonstrate that it is in principle possible to compute non linear
input manifolds that map to affine output manifolds. This demonstrates
the possibility of deep convolutional networks to achieve flattening
of input data which is generally considered as an important
preprocessing step for classification. Since we do not consider a
complete network with fully connected layers at the end we cannot give
details how classification is achieved. The explicit demonstration of
non linear manifolds that map to affine outputs however indicates a
possible basic structure of input manifolds for classes. It is easy to
see that a parallel translation of the affine output manifold would
result in two linearly separable manifolds that would be generated by
essentially parallel translated non linear manifolds in the input
space. This demonstrates that convolutional networks can be designed
to exactly separate sufficiently ``covariant `` classes.  and that this
could be the reason for the relative success of convolutional networks
over previous machine learning approaches to classification and explain 
why using a large number of classes for training is advantageous since
they all contribute to very similar individual manifolds.

Disregarding these speculations the fact remains that these manifolds
will always exist since they are derived on purely formal grounds from
the structure of the network. If they have no role in classification
their presence will have to be explained in other ways.



\bibliography{stefan}
\bibliographystyle{plain}


\end{document}